\documentclass{article}

\usepackage[preprint]{neurips_2023}

\usepackage[utf8]{inputenc} 
\usepackage[T1]{fontenc}    
\usepackage{hyperref}       
\usepackage{url}            
\usepackage{booktabs}       
\usepackage{amsfonts}       
\usepackage{nicefrac}       
\usepackage{microtype}      
\usepackage{xcolor}         

\usepackage{xspace} 
\usepackage[T1]{fontenc} 
\usepackage{amsmath}
\usepackage[capitalize]{cleveref}
\usepackage{algorithm}
\usepackage{algpseudocode}
\usepackage{bm}
\usepackage{graphicx}
\usepackage{graphicx}
\usepackage{subcaption}
\usepackage{tabularray}
\usepackage{booktabs}
\usepackage{numprint}

\makeatletter
\DeclareRobustCommand\onedot{\futurelet\@let@token\@onedot}
\def\@onedot{\ifx\@let@token.\else.\null\fi\xspace}

\def\eg{{e.g}\onedot} 
\def\ie{{i.e}\onedot} 
\def\aka{{a.k.a}\onedot}

\makeatother

\newcommand{\Sz}{\mathcal{S}_{\vec{\phi}}(\vec{z})}

\newcommand{\invS}{\mathcal{S}^{-1}_{\vec{\phi}}(\vec{\varepsilon})}
\newcommand{\SNormal}{\mathcal{S}_{\mu, \sigma}(z)}

\renewcommand{\vec}[1]{\bm{#1}}

\newcommand{\E}[2]{\mathop{\mathbb{E}}_{#1}\left[#2\right]}

\DeclareMathOperator{\softplus}{softplus}

\hypersetup{pdfauthor={Luca Della Libera},
pdftitle={Soft Actor-Critic with Beta Policy via Implicit Reparameterization Gradients},
pdfsubject={Deep reinforcement learning},
pdfkeywords={Soft actor-critic, beta distribution, implicit reparameterization gradients, MuJoCo, deep reinforcement learning}}

\usepackage{enumitem}
\setlist[itemize]{leftmargin=7mm}

\title{Soft Actor-Critic with Beta Policy via Implicit Reparameterization Gradients}

\author{%
  Luca Della Libera \\
  Gina Cody School of Engineering and Computer Science \\
  Concordia University\\
  \texttt{luca.dellalibera@mail.concordia.ca} \\
}

\begin{document}

\maketitle

\begin{abstract}
Recent advances in deep reinforcement learning have achieved impressive results in a wide range of complex tasks, but poor sample efficiency remains a major obstacle to real-world deployment.
Soft actor-critic (SAC) mitigates this problem by combining stochastic policy optimization and off-policy learning, but its applicability is restricted to distributions whose gradients can be computed through the reparameterization trick. This limitation excludes several important examples such as the beta distribution, which was shown to improve the convergence rate of actor-critic algorithms in high-dimensional continuous control problems thanks to its bounded support.
To address this issue, we investigate the use of implicit reparameterization, a powerful technique that extends the class of reparameterizable distributions. In particular, we use implicit reparameterization gradients to train SAC with the beta policy on simulated robot locomotion environments and compare its performance with common baselines. 
Experimental results show that the beta policy is a viable alternative, as it outperforms the normal policy and is on par with the squashed normal policy, which is the go-to choice for SAC. The code is available at \href{https://github.com/lucadellalib/sac-beta}{https://github.com/lucadellalib/sac-beta}.
\end{abstract}

\section{Introduction}
\label{sec:introduction}
In recent years, we have witnessed impressive advances in deep reinforcement learning, with successful applications on a wide range of complex tasks, from playing games~\cite{atari_mnih, Silver_2016} to high-dimensional continuous control~\cite{gae_schulman}. However, the poor sample efficiency of deep reinforcement learning is a major obstacle to its use in real-world domains, especially in tasks such as robotic manipulation that require interaction without the safety net of simulation. The high risk of damaging and the costs associated with physical robot experimentation make it crucial for the learning process to be sample efficient. As a result, it is of paramount importance to develop algorithms that can learn from limited data and generalize effectively to new situations.
To address this issue, \cite{sac_haarnoja} introduced soft actor-critic (SAC), an off\nobreakdash-policy model\nobreakdash-free algorithm that bridges the gap between stochastic policy optimization and off\nobreakdash-policy learning, achieving state\nobreakdash-of\nobreakdash-the\nobreakdash-art on the popular  multi-joint dynamics with contact (MuJoCo)~\cite{mujoco} benchmark suite.
Based on the maximum entropy framework, SAC aims to maximize a trade\nobreakdash-off between the expected return and the entropy of the policy, \ie to accomplish the task while acting as randomly as possible. As a consequence, the agent is encouraged to explore more widely, and to assign identical probability mass to actions that are equally attractive.
A key step of the algorithm, necessary for computing the entropy\nobreakdash-augmented objective function, is sampling an action from the current policy in a differentiable manner, which is accomplished through the use of the reparameterization trick. However, certain distributions such as gamma, beta, Dirichlet and von Mises cannot be used with the reparameterization trick. Hence, the applicability of SAC to practical problems that could benefit from the injection of prior knowledge through an appropriate distribution family is limited. This restriction is particularly relevant for domains where the choice of the distribution is critical to achieving good performance, such as tasks involving actions with specific bounds or constraints.

Normalizing flows~\cite{rezende_flow,dinh_nf} are an attractive solution to this problem. By using a series of invertible transformations to map a simple density to a more complex one via the change of variable formula, normalizing flows improve the expressiveness of the policy while retaining desirable properties such as the ease of reparameterization.
For example, the $\tanh$ squashing function originally used in SAC~\cite{sac_haarnoja} to enforce action bounds can be interpreted as a single layer normalizing flow with no learnable parameters.
\cite{normalizing-flows_mazoure} further extend it by incorporating radial, planar and autoregressive flows to allow for a broader range of distributions (\eg multimodal), resulting in improved performance on variety of tasks.
A similar study by \cite{normalizing-flows_ward} shows that combining SAC with modified Real NVP flows~\cite{dinh_nf} leads to better stability and increased exploration capabilities in sparse reward environments.
While normalizing flows have many useful properties, there are also some drawbacks to their use in deep reinforcement learning. One issue is that they can be computationally demanding for large-scale problems or when dealing with high-dimensional inputs. Additionally, they can be difficult to train, especially when using complex flow architectures.

An orthogonal approach to normalizing flows that addresses the problem of differentiating through stochastic nodes is implicit reparameterization~\cite{ad,omt}, which enables the computation of gradients for a broader class of distribution families for which the reparameterization trick is not applicable.
In this work, we use implicit reparameterization gradients to train SAC with the beta policy, which is bounded to the unit interval and was shown to substantially improve the convergence rate of actor-critic algorithms such as TRPO~\cite{trpo_schulman} and ACER~\cite{acer_wang} (which do not require differentiating through random variables)~\cite{beta_chou} on continuous control problems. In particular, we explore two variants of implicit reparameterization and we evaluate their performance on four MuJoCo~\cite{mujoco} environments. Experiments demonstrate that the beta policy outperforms the normal policy and yields similar results to the squashed normal policy~\cite{sac_haarnoja}, which is the go-to choice for SAC.

\section{Background}
\label{sec:background}

\subsection{Markov Decision Process}\label{sec:rl_problem}
A reinforcement learning problem~\cite{rl_sutton} can be formalized as a Markov decision process, defined by a tuple $(\mathcal{S}, \mathcal{A}, \mathcal{R}, \mathcal{P})$, where $\mathcal{S}$ denotes the {state space}, $\mathcal{A}$ the {action space}, $\mathcal{R}:\ \mathcal{S} \times \mathcal{A} \rightarrow \mathbb{R}$ the reward function and $\mathcal{P}:\ \mathcal{S} \times \mathcal{A} \rightarrow [0, \infty)$ the transition probability density function. At each time step $t$, the agent observes a state $\vec{s}_t \in \mathcal{S}$ and performs an action $\vec{a}_t \in \mathcal{A}$ according to a policy $\pi:\ \mathcal{S} \times \mathcal{A} \rightarrow [0, \infty)$, which results in a reward $r_{t+1} = \mathcal{R}(\vec{s}_t, \vec{a}_t)$ and a next state $\vec{s}_{t+1} \sim \mathcal{P}(\vec{s}_t, \vec{a}_t)$.
The {return} associated with a {trajectory} (\aka episode or {rollout}) $\vec{\tau} = (\vec{s}_0, \vec{a}_0, r_1, \vec{s}_1, \vec{a}_1, r_2, \, \dots)$ is defined as
\begin{equation}
R(\vec{\tau}) = \sum_{t=0}^{\infty} \gamma^t r_{t+1},
\end{equation}
\ie the {infinite\nobreakdash-horizon} cumulative discounted reward with discount factor $\gamma \in (0,1)$.
The agent aims to maximize the expected return
\begin{equation}
\label{eq:return}
J(\pi) = \E{\vec{\tau} \sim \pi}{R(\vec{\tau})} = \E{\vec{\tau} \sim \pi}{\sum_{t=0}^{\infty} \gamma^t r_{t+1}},
\end{equation}
\ie to learn an optimal policy $\pi^\star = \arg \max_{\pi} J(\pi)$. The {state value function} (\aka state value) $V^{\pi}(\vec{s})$ of a state $\vec{s}$ is defined as
\begin{equation}\label{eq:v_value}
    V^{\pi}(\vec{s}) = \E{\vec{\tau} \sim \pi}{R(\vec{\tau}) \mid \vec{s}_0 = \vec{s}},
\end{equation}
\ie the expected return obtained by the agent if it starts in state $\vec{s}$ and always acts according to policy $\pi$.
Similarly, the {state\nobreakdash-action value function} (\aka action value or {Q\nobreakdash-value}) $Q^{\pi}(\vec{s},\vec{a})$ of a state\nobreakdash-action pair $(\vec{s}, \vec{a})$ is defined as
\begin{equation}\label{eq:q_value}
    Q^{\pi}(\vec{s}, \vec{a}) = \E{\vec{\tau} \sim \pi}{R(\vec{\tau}) \mid \vec{s}_0 = \vec{s}, \, \vec{a}_0 = \vec{a}},
\end{equation}
\ie the expected return obtained by the agent if it starts in state $\vec{s}$, performs action $\vec{a}$ and always acts according to policy $\pi$.

\subsection{Soft Actor-Critic}
\label{subsec:sac}
SAC~\cite{sac_haarnoja} redefines the expected return in \cref{eq:return} to be maximized by policy $\pi$ as
\begin{equation}
J(\pi) = \E{\vec{\tau} \sim \pi}{\sum_{t=0}^{\infty} \gamma^t (r_{t + 1} + \alpha H(\pi(\, \cdot \mid \vec{s}_t)))},
\end{equation}
where $\vec{\tau} = (\vec{s}_0, \vec{a}_0, r_1, \vec{s}_1, \vec{a}_1, r_2, \, \dots)$ is a trajectory of state\nobreakdash-action\nobreakdash-reward triplets sampled from the environment, $\gamma$ the discount factor, $H(\pi(\, \cdot \mid \vec{s}_t)) = \E{\vec{a} \sim \pi(\vec{a} \mid \vec{s}_t)}{-\log \pi(\vec{a} \mid \vec{s}_t)}$ the entropy of $\pi$ in state $\vec{s}_t$, and $\alpha > 0$ a temperature parameter that controls the relative strength of the entropy regularization term.
Since $J(\pi)$ now includes a state\nobreakdash-dependent entropy bonus, the definitions of $V^{\pi}(\vec{s})$ and $Q^{\pi}(\vec{s},\vec{a})$ need to be updated accordingly as
\begin{equation}
V^{\pi}(\vec{s}) = \E{\vec{\tau} \sim \pi}{\sum_{t=0}^{\infty} \gamma^t (r_{t+1} + \alpha H(\pi(\, \cdot \mid \vec{s}_t)) \,\middle\vert\, \vec{s}_0 = \vec{s}},
\end{equation}
\begin{equation}
Q^{\pi}(\vec{s},\vec{a}) = \E{\vec{\tau} \sim \pi}{\sum_{t=0}^{\infty} \gamma^t r_{t+1} + \alpha \sum_{t=1}^{\infty} \gamma^t H(\pi(\, \cdot \mid \vec{s}_t)) \,\middle\vert\, \vec{s}_0 = \vec{s}, \, \vec{a}_0 = \vec{a}}.
\end{equation}
Note that the modified $Q^{\pi}(\vec{s},\vec{a})$ does not include the entropy bonus from the first time step.
Let $Q_{\vec{\psi}}(\vec{s},\vec{a})$ be the action value function approximated by a neural network with weights $\vec{\psi}$ and $\pi_{\vec{\theta}}$ a stochastic policy parameterized by a neural network with weights $\vec{\theta}$.
The action value network is trained to minimize the squared difference between $Q_{\vec{\psi}}(\vec{s},\vec{a})$ and the {soft} temporal difference estimate of $Q^{\pi}(\vec{s},\vec{a})$, defined as
\begin{equation}
    \widehat{Q}(\vec{s}_t,\vec{a}_t) = r_{t + 1} + \gamma (Q_{\vec{\psi}}(\vec{s}_{t+1},\tilde{\vec{a}}_{t+1}) - \alpha \log \pi_{\vec{\theta}}(\tilde{\vec{a}}_{t+1} \mid \vec{s}_{t+1})),
\end{equation}
which is the standard temporal difference estimate~\cite{td_sutton} augmented with the entropy bonus, where $r_{t + 1}$ and $\vec{s}_{t+1}$ are sampled from a replay buffer storing past experiences and $\tilde{\vec{a}}_{t+1}$ is sampled from the {current} policy $\pi_{\vec{\theta}}(\, \cdot \mid \vec{s}_{t+1})$.
In practice, the approximation $Q_{\vec{\psi}}(\vec{s}_{t+1},\vec{a}_{t+1})$ is computed by a {target network}~\cite{human_mnih}, whose weights $\vec{\psi}_{\text{target}}$ are periodically updated as $\vec{\psi}_{\text{target}} \leftarrow (1 - \tau) \vec{\psi}_{\text{target}} + \tau \vec{\psi}$, with smoothing coefficient $\tau \in [0, 1]$.
Not only does this help to stabilize the training process, but it also turns the ill\nobreakdash-posed problem of learning $Q_{\vec{\psi}}(\vec{s},\vec{a})$ via bootstrapping into a supervised learning one that can be solved via gradient descent.
Furthermore, double Q\nobreakdash-learning is used to reduce the overestimation bias and speed up convergence~\cite{double-dqn_van-hasselt}. This means that $Q_{\vec{\psi}}(\vec{s},\vec{a})$ is computed as the minimum between two action value function approximations $Q_{\vec{\psi}_1}(\vec{s},\vec{a})$ and $Q_{\vec{\psi}_2}(\vec{s},\vec{a})$, parameterized by neural networks with weights $\vec{\psi}_1$ and $\vec{\psi}_2$, respectively.

While the action value network learns to minimize the error in the Q\nobreakdash-value approximation, $\pi_{\vec{\theta}}$ is updated via gradient ascent to maximize
\begin{equation}
J(\pi_{\vec{\theta}}) = \E{\substack{\vec{a} \sim \pi_{\vec{\theta}} \\ \vec{s} \sim \mathcal{P}}}{Q_{\vec{\psi}}(\vec{s},\vec{a}) - \alpha \log \pi_{\vec{\theta}}({\vec{a}} \mid \vec{s}))},
\end{equation}
where $\mathcal{P}$ is the transition probability function, $Q_{\vec{\psi}}(\vec{s},\vec{a}) = \min \{Q_{\vec{\psi}_1}(\vec{s},\vec{a}), Q_{\vec{\psi}_2}(\vec{s},\vec{a})\}$, and $\vec{a}$ is drawn from $\pi_{\vec{\theta}}$ in a {differentiable} way through the reparameterization trick.
The temperature parameter $\alpha$, which explicitly handles the exploration\nobreakdash-exploitation trade\nobreakdash-off, is particularly sensitive to the magnitude of the reward, and needs to be tuned manually for each task. 
The final algorithm, which alternates between sampling transitions from the environment and updating the neural networks using the experiences retrieved from a replay buffer, is shown in \cref{alg:sac}.
Note that, although in theory the replay buffer can store an arbitrary number of transitions, in practice a maximum size should be specified based on the available memory. When the replay buffer overflows, past experiences are removed according to the First\nobreakdash-In First\nobreakdash-Out (FIFO) rule. It is also advisable to collect a minimum number of transitions prior to the start of the training process to provide the neural networks with a sufficient amount of uncorrelated experiences to learn from.

\begin{algorithm}[t]
\small
\caption{Soft Actor-Critic~\cite{sac_haarnoja}}
\label{alg:sac}
\begin{algorithmic}
\Require initial $\vec{\theta}$, $\vec{\psi}_1$, $\vec{\psi}_2$
\State {Initialize} weights of the target networks
\State {Initialize} empty replay buffer
\For{each iteration}
	\For{each time step}
	    \State {Sample} action from the policy
	    \State {Sample} transition from the environment
	    \State {Store} transition into the replay buffer
	\EndFor
	\For{each update step}
	    \State {Update} weights of the action value networks
	    \State {Update} weights of the policy network
	    \State {Update} weights of the target networks
    \EndFor
\EndFor
\Ensure optimized $\vec{\theta}$, $\vec{\psi}_1$, $\vec{\psi}_2$
\end{algorithmic}
\end{algorithm}

\subsection{Implicit Reparameterization Gradients}
\label{subsec:implicit_gradient_reparameterization}
Implicit reparameterization~\cite{ad,omt} is an alternative to the reparameterization trick (\aka explicit reparameterization) that enables the computation of gradients of stochastic computations for a broader class of distribution families.
Explicit reparameterization requires an invertible and continuously differentiable standardization function $\Sz$ that, when applied to a sample from distribution $q_{\vec{\phi}} (\vec{z})$, removes its dependency on the distribution parameters $\vec{\phi}$:
\begin{equation}
    \Sz = \vec{\varepsilon} \sim q(\vec{\varepsilon}) \qquad \vec{z} = \invS. \label{eqn:z-g-inverse}
\end{equation}
For instance, in the case of a normal distribution $\mathcal{N}(\mu, \sigma)$ a valid standardization function is $\SNormal = (z - \mu)/\sigma \sim \mathcal{N}(0, 1)$. Let $f(\vec{z})$ denote an objective function whose expected value over $q_{\vec{\phi}}$ is to be optimized with respect to $\vec{\phi}$. Then
\begin{equation}
\E{q_{\vec{\phi}} (\vec{z})} {f(\vec{z})} = \E{q(\vec{\varepsilon})}{f(\invS)}
\end{equation}
and the gradient of the expectation can be computed via chain rule as
\begin{equation}
    \nabla_{\vec{\phi}} \E{q_{\vec{\phi}} (\vec{z})} {f(\vec{z})} = \E{q(\vec{\varepsilon})} {\nabla_{\vec{\phi}} f(\invS)} = \E{q(\vec{\varepsilon})} {\nabla_{\vec{z}} f(\invS) \nabla_{\vec{\phi}} \invS}.
    \label{eqn:explicit-reparameterization}
\end{equation}
Although many continuous distributions admit a standardization function, it is often non\nobreakdash-invertible or prohibitively expensive to invert, hence the reparameterization trick is not applicable. This is the case for distributions such as gamma, beta, Dirichlet and von Mises. Fortunately, implicit reparameterization relaxes this constraint, allowing for gradient computations without $\invS$.
Using the change of variable $\vec{z} = \invS$, \cref{eqn:explicit-reparameterization} can be rewritten as:
\begin{equation}
    \nabla_{\vec{\phi}} \E{q_{\vec{\phi}} (\vec{z})} {f(\vec{z})} = \E{q_{\vec{\phi}}{(\vec{z})}}{\nabla_{\vec{z}} f(\vec{z}) \nabla_{\vec{\phi}} \vec{z}}, \qquad \nabla_{\vec{\phi}} \vec{z} = \nabla_{\vec{\phi}} \invS |_{\vec{\varepsilon} = \Sz}.
    \label{eq:explicit-grad-z}
\end{equation}
At this point, $\nabla_{\vec{\phi}} \vec{z}$ can be obtained via implicit differentiation by applying the total gradient operator to the equation $\Sz = \vec{\varepsilon}$:
\begin{equation}
    \nabla_{\vec{\phi}} \Sz = \nabla_{\vec{\phi}} \vec{\varepsilon} \iff
    \nabla_{\vec{z}} \Sz \nabla_{\vec{\phi}} \vec{z} + \nabla_{\vec{\phi}} \Sz = \vec{0} \iff 
    \boxed{ \nabla_{\vec{\phi}} \vec{z} = -(\nabla_{\vec{z}} \Sz)^{-1} \nabla_{\vec{\phi}} \Sz }
    \label{eqn:z-grad}
\end{equation}
Remarkably, this expression for the gradient does not require inverting the standardization function but only differentiating it.

\section{Method}
\label{sec:beta}
The main goal of this work is to explore the use of SAC in combination with the beta policy. To do so, we employ implicit reparameterization to enable differentiation through the sampling process. As explained in \cref{subsec:implicit_gradient_reparameterization}, this approach only requires backpropagating through the standardization function $\Sz$ without inverting it.
Hence, we need to define an appropriate $\Sz$ for the beta distribution and derive an expression for $\nabla_{\vec{\phi}} \vec{z}$. Following \cite{ad}, $z \sim \operatorname{Beta}(\alpha, \beta)$ can also be obtained as $z = \frac{z_1}{z_1 + z_2}$, where $z_1 \sim \operatorname{Gamma}(\alpha, 1)$ and $z_2 \sim \operatorname{Gamma}(\beta, 1)$. Then the problem reduces to calculating implicit reparameterization gradients for the gamma distribution $\operatorname{Gamma}(\alpha, \beta)$. Being its density closed under scaling transforms, a valid standardization function for $\beta$ is simply $\mathcal{S}_{\beta}(z) = \frac{z}{\beta}$. However, for $\alpha$, $\mathcal{S}_{\alpha}(z)$ is the regularized incomplete gamma function, which cannot be expressed analytically. Therefore, it is necessary to resort to approximations. A possible solution is the one proposed by \cite{ad}, who apply forward-mode automatic differentiation to algorithm AS 32 \cite{as_gamma}, a numerical method that approximates its value.
An alternative approach by \cite{omt}, inspired by the theory of optimal mass transport, derives instead a minimax closed\nobreakdash-form approximation based on Taylor expansion of the derivative of the beta cumulative distribution function, which is also a valid standardization function. Nevertheless, as reported by~\cite{ad}, it is slower and less accurate than the automatic differentiation strategy.
In this study, we experiment with both techniques and we compare the resulting implicitly reparameterized beta policy to more frequently used ones such as normal and squashed normal.
In summary, we consider the following four SAC variants:
\begin{itemize}
\item \textbf{SAC-Beta-AD}: SAC with a beta policy whose gradient is computed via automatic differentiation implicit reparameterization~\cite{ad}.
\item \textbf{SAC-Beta-OMT}: SAC with a beta policy whose gradient is computed via optimal mass transport implicit reparameterization~\cite{omt}.
\item \textbf{SAC-Normal}: SAC with a normal policy whose gradient is computed via reparameterization trick.
\item \textbf{SAC-TanhNormal}: SAC with a squashed normal policy, obtained by applying $\tanh$ to samples drawn from a normal distribution~\cite{sac_haarnoja}, whose gradient is computed via reparameterization trick.
\end{itemize}

\section{Experiments}
\label{sec:mujoco}
\subsection{MuJoCo Environments}
We evaluate the proposed methods on four of the eleven MuJoCo environments, namely Ant-v4, HalfCheetah-v4, Hopper-v4 and Walker2d-v4 (see \cref{fig:mujoco}). The goal is to walk as fast as possible without falling while at the same time minimizing the number of actions and reducing the impact on each joint. Observations consist of coordinate values of different robot's body parts, followed by the velocities of each of those individual parts. Actions, bounded to $[-1, 1]$, represent the torques applied at the joints. The dimensionality of both the observation and the action space varies across the environments, depending on the complexity of the task (see \cref{tab:environments}).

\subsection{Training Setup and Hyperparameters}
Hyperparameter values are reported in \cref{tab:hyperparameters}.
We use the same architecture for both the policy and the action value network, consisting of $2$ fully connected hidden layers with ReLU nonlinearities. The action value network receives as an input the concatenation of the flattened observation and action and returns a Q\nobreakdash-value. The policy network receives as an input the flattened observation and returns the parameters of the corresponding action distribution. In particular, for SAC\nobreakdash-Beta\nobreakdash-AD and SAC\nobreakdash-Beta\nobreakdash-OMT, it returns the logarithm of the concentration parameters shifted by $1$ to ensure that the distribution is both concave and unimodal~\cite{beta_chou}. For SAC\nobreakdash-Normal and SAC\nobreakdash-TanhNormal, it returns the mean and the logarithm of the standard deviation. For all distributions, we assume the action dimensions are independent.
To improve training stability, we clip the log shifted concentrations and log standard deviation to the interval $[-20, 2]$.
Furthermore, we clip the samples drawn from the beta distribution to the interval $[10^{-7}, 1 - 10^{-7}]$ to avoid underflow and we linearly map them from $[0, 1]$ to the environment's action interval $[-1, 1]$.
We train the policy for $10^6$ time steps using Adam optimizer~\cite{adam} with a learning rate of $0.001$ and a batch size of $256$. We perform $1$ gradient update per time step and synchronize the target networks after each update with smoothing coefficient $0.005$. We use a discount factor of $0.99$, a temperature of $0.2$ and a replay buffer of size $10^6$, which is filled with $10^4$ initial transitions collected by sampling actions uniformly at random.
We test the policy every $5000$ time steps for $10$ episodes using the distribution mean as the predicted action and we report the average return.

\begin{table}[t]
    \centering
    {\begin{tabular}{c|c|c}
        \textbf{Environment} & \textbf{Observation dimensions}  & \textbf{Action dimensions}    \\
    \hline \hline
     Ant-v4 &  $27$ & $8$ \\
     HalfCheetah-v4 & $17$ & $6$   \\
     Hopper-v4 & $11$ & $3$ \\
     Walker2d-v4 & $17$ & $6$ \\
         \hline
    \end{tabular}}
    \vspace{.5em}
    \caption{Observation and action dimensions of the four considered MuJoCo environments.}
    \label{tab:environments}
\end{table}

\begin{table}[t]
    \centering
    {\begin{tabular}{l|l}
    \textbf{Parameter} & \textbf{Value} \\
    \hline \hline
     Number of hidden layers (all networks) & $2$\\
     Number of neurons per hidden layer & $256$\\
     Nonlinearity & ReLU\\
     Optimizer & Adam \cite{adam}\\
     Learning rate & $0.001$\\
     Batch size & $256$\\
     Replay buffer size & $10^6$\\
     Discount factor &  $0.99$\\
     Target smoothing coefficient & $0.005$\\
     Target update frequency & $1$\\
     Temperature & $0.2$\\
     Log standard deviation interval & $[-20, 2]$\\
     Log shifted concentration interval & $[-20, 2]$\\
     Beta sample interval & $[10^{-7}, 1 - 10^{-7}]$\\
     Number of gradient updates per time step & $1$\\
     Number of initial time steps & $10^4$\\
     Total number of time steps & $10^6$\\
     Test frequency & $5000$ \\
     Number of test episodes & $10$ \\
     \hline
    \end{tabular}}
    \vspace{.5em}
    \caption{SAC hyperparameters.}
    \label{tab:hyperparameters}
\end{table}

\subsection{Implementation and Hardware}
Software for the experimental evaluation was implemented in Python 3.10.2 using Gymnasium 0.28.1\footnote{\href{https://github.com/Farama-Foundation/Gymnasium/tree/v0.28.1}{https://github.com/Farama-Foundation/Gymnasium/tree/v0.28.1}}~\cite{gym_brockman} + EnvPool 0.8.1\footnote{\href{https://github.com/sail-sg/envpool/tree/v0.8.1}{https://github.com/sail-sg/envpool/tree/v0.8.1}}~\cite{envpool} for the MuJoCo environments, Tianshou 0.5.0\footnote{\href{https://github.com/thu-ml/tianshou/tree/v0.5.0}{https://github.com/thu-ml/tianshou/tree/v0.5.0}}~\cite{tianshou} for SAC, PyTorch 1.13.1\footnote{\href{https://github.com/pytorch/pytorch/tree/v1.13.1}{https://github.com/pytorch/pytorch/tree/v1.13.1}}~\cite{pytorch_paszke} for the neural network architectures, probability distributions and reference implementation of optimal mass transport implicit reparameterization, TensorFlow 2.11.0\footnote{\href{https://github.com/tensorflow/tensorflow/tree/v2.11.0}{https://github.com/tensorflow/tensorflow/tree/v2.11.0}}~\cite{tensorflow} + TensorFlow Probability 0.19.0\footnote{\href{https://github.com/tensorflow/probability/tree/v0.19.0}{https://github.com/tensorflow/probability/tree/v0.19.0}}~\cite{tfp} for the reference implementation of automatic differentation implicit reparameterization and Matplotlib 3.7.0\footnote{\href{https://github.com/matplotlib/matplotlib/tree/v3.7.0}{https://github.com/matplotlib/matplotlib/tree/v3.7.0}}~\cite{hunter_matplotlib} for plotting.
All the experiments were run on a CentOS Linux 7 machine with an Intel Xeon Gold 6148 Skylake CPU with $20$ cores @ $2.40$ GHz, $32$ GB RAM and an NVIDIA Tesla V100 SXM2 @ $16$ GB with CUDA Toolkit 11.4.2.

\subsection{Results and Discussion}
\begin{figure}[t]
    \hspace{-2.5mm}
    \begin{subfigure}{.245\textwidth}
      \includegraphics[width=1.0\linewidth]{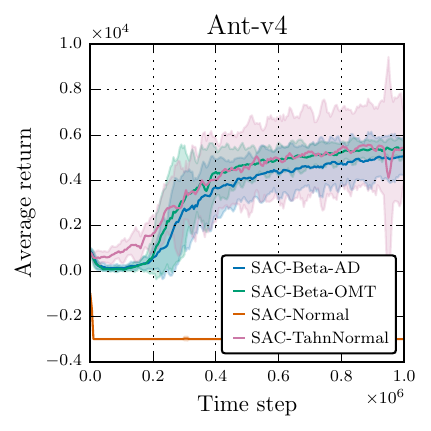}
    \end{subfigure}
    \hspace{-1mm}
    \begin{subfigure}{.245\textwidth}
      \includegraphics[width=1.0\linewidth]{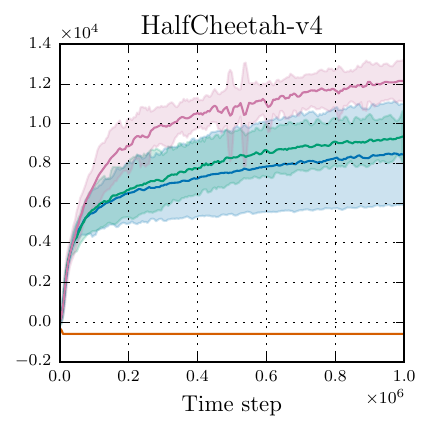}
    \end{subfigure}
    \hspace{-1mm}
    \begin{subfigure}{.245\textwidth}
      \includegraphics[width=1.0\linewidth]{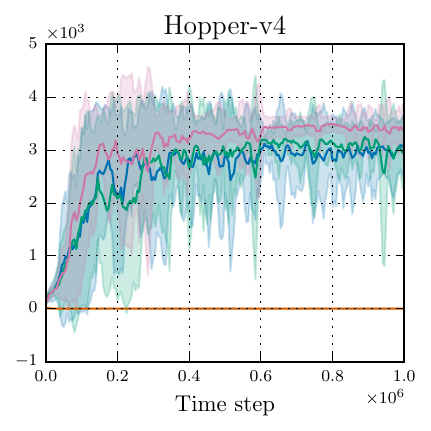}
    \end{subfigure}
    \begin{subfigure}{.245\textwidth}
      \includegraphics[width=1.0\linewidth]{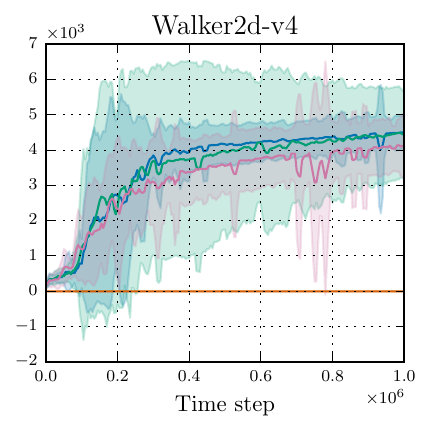}
    \end{subfigure}
    \centering
    \begin{subfigure}{.20\textwidth}
      \includegraphics[width=1.0\linewidth]{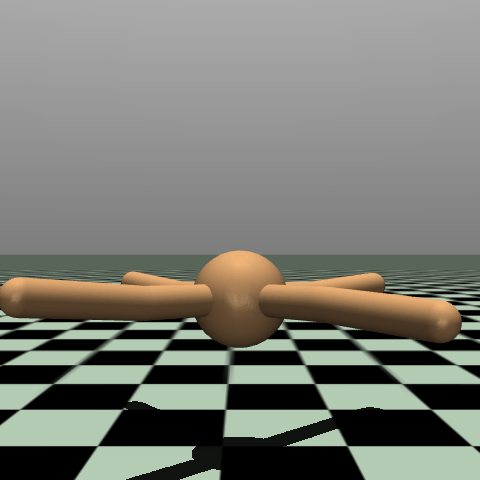}
    \end{subfigure}
    \hspace{4mm}
    \begin{subfigure}{.20\textwidth}
      \includegraphics[width=1.0\linewidth]{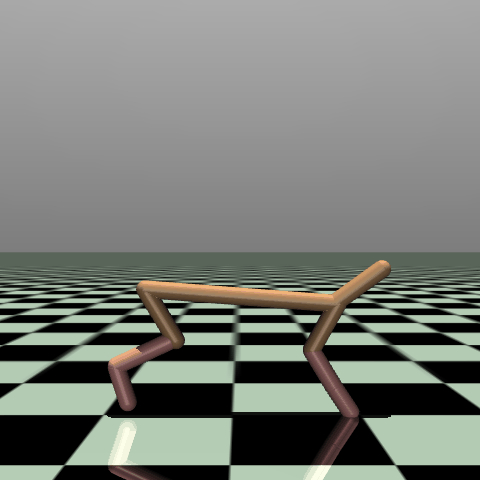}
    \end{subfigure}
    \hspace{4mm}
    \begin{subfigure}{.20\textwidth}
      \includegraphics[width=1.0\linewidth]{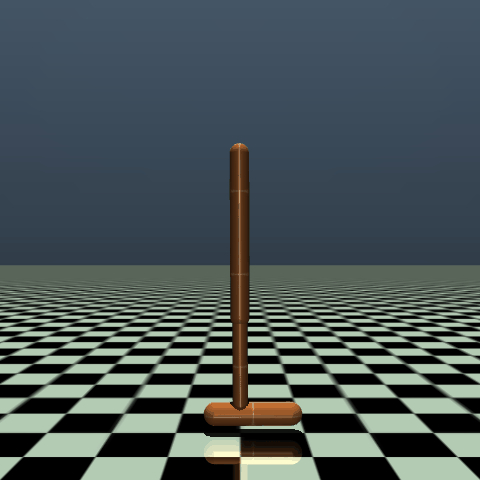}
    \end{subfigure}
    \hspace{4mm}
    \begin{subfigure}{.20\textwidth}
      \includegraphics[width=1.0\linewidth]{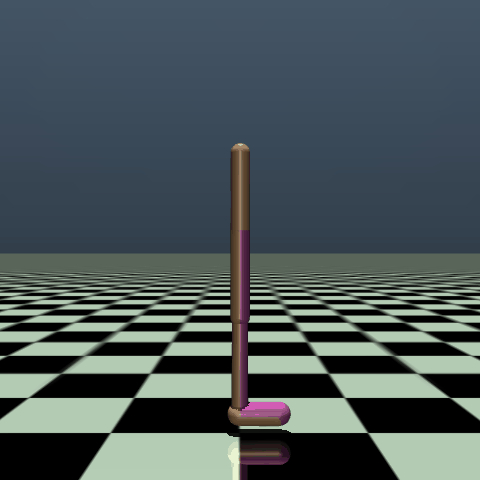}
    \end{subfigure}
\caption{
Performance comparison on the four considered MuJoCo environments. Curves represent the average return $\pm$ $2$ standard deviations over $5$ random seeds, smoothed using a centered moving average with radius $1$.}
\label{fig:mujoco}
\end{figure}

A comparison between the proposed SAC variants is presented in \cref{fig:mujoco}, with the final performance reported in \cref{tab:reported_rewards}. We observe that SAC-Normal struggles to learn anything useful, as the average return remains close to zero in all the environments. This could be ascribed to a poor initialization, as the training process immediately diverges and the policy gets stuck in a region of the search space from which it cannot recover.
Further investigations are needed to understand why the normal policy performs poorly with SAC while it does not exhibit the same flawed behavior with other non\nobreakdash-entropy\nobreakdash-based algorithms such as TRPO~\cite{trpo_schulman,beta_chou}.
Overall, SAC-Beta-AD and SAC-Beta-OMT perform similarly to SAC-TanhNormal, with slightly higher final return in Ant\nobreakdash-v4 and Walker2d-v4 but worse in HalfCheetah-v4 and Hopper-v4. However, the large standard deviation values indicate that more experiments might be necessary to obtain more accurate estimates.
Regarding the comparison between SAC-Beta-AD and SAC-Beta-OMT, we only observe little differences mostly due to random fluctuations. This is surprising, since we were expecting faster convergence with SAC-Beta-AD, given the fact that it provides more accurate gradient estimates~\cite{ad}. This suggests that highly accurate gradients may not be critical to the algorithm's success. Therefore, a simpler gradient estimator such as the score function estimator~\cite{estimators}, which does not rely on assumptions about the distribution family, could be a promising alternative to explore.

\begin{table}[t]
    \centering
    {\begin{tabular}{c|c|c|c|c}
    \textbf{Environment} & \textbf{SAC-Beta-AD} & \textbf{SAC-Beta-OMT} & \textbf{SAC-Normal} & \textbf{SAC-TanhNormal} \\
    \hline \hline
     Ant-v4 & $\numprint{5068} \pm \numprint{940}$ & $\mathbf{\numprint{5456}} \pm \mathbf{\numprint{260}}$  & $\numprint{-3000} \pm 3$ & ${\numprint{5309}} \pm {\numprint{1324}}$  \\
     HalfCheetah-v4& $\numprint{8363} \pm \numprint{2469}$ & $\numprint{9378} \pm \numprint{1342}$  & $\numprint{-599} \pm 3$ & $\mathbf{\numprint{12198}} \pm \mathbf{\numprint{1054}}$  \\
     Hopper-v4& $\numprint{2946} \pm \numprint{956}$ & $\numprint{2946} \pm \numprint{806}$  & $4 \pm 2$ & $\mathbf{\numprint{3316}} \pm \mathbf{\numprint{767}}$ \\
     Walker2d-v4  & $\mathbf{\numprint{4523}} \pm \mathbf{\numprint{409}}$ & $\numprint{4420} \pm \numprint{1230}$  & $-7 \pm 2$ & $\numprint{4198} \pm \numprint{882}$ \\
         \hline
    \end{tabular}}
    \vspace{.5em}
    \caption{Final average return $\pm$ $2$ standard deviations over $5$ random seeds on the four considered MuJoCo environments.}
    \label{tab:reported_rewards}
\end{table}

\begin{figure}[t]
    \centering
    \includegraphics[width=0.5\linewidth]{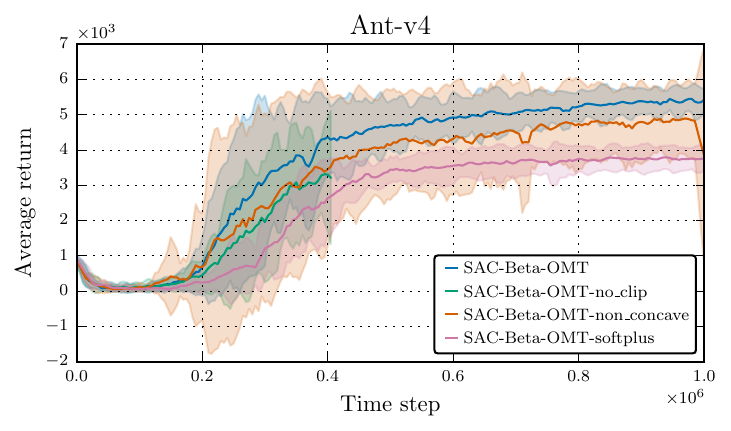}
    \caption{
Ablation study for SAC-Beta-OMT on Ant-v4. Curves represent the average return $\pm$ $2$ standard deviations over $5$ random seeds, smoothed using a centered moving average with radius $1$.}
    \label{fig:ablation}
\end{figure}

We also conduct an ablation study for SAC-Beta-OMT on Ant-v4 to gain more insights on the impact of our design choices. In particular, we consider the following three ablations:
\begin{itemize}
\item \textbf{SAC-Beta-OMT-no\_clip}: we do not clip the log shifted concentrations.
\item \textbf{SAC-Beta-OMT-non\_concave}: we do not shift the concentrations, allowing for non\nobreakdash-concave beta distributions.
\item \textbf{SAC-Beta-OMT-softplus}: following \cite{beta_chou}, we model the concentrations using $\softplus$ instead of $\exp$, meaning that the policy network effectively returns inverse softplus shifted concentrations.
\end{itemize}
As depicted in \cref{fig:ablation}, SAC-Beta-OMT performs the best, demonstrating the effectiveness of our training setup. Note that SAC-Beta-OMT-no\_clip terminates prematurely after $\sim 400$ k time steps due to exploding or vanishing concentrations, leading us to the conclusion that clipping is crucial to avoid instabilities in the optimization process.

\section{Conclusion}
\label{sec:conclusion}
In this work, we utilized implicit reparameterization gradients~\cite{ad,omt} to train SAC with the beta policy, which is an alternative to normalizing flows in addressing the bias issue resulting from the mismatch between the infinite support of the normal distribution and the bounded action space typically present in real-world settings.
Experiments on four MuJoCo environments show that the beta policy is a viable alternative, as it outperforms the normal policy and yields similar results to the squashed normal policy, frequently used together with SAC.
Future research includes analyzing the qualitative behavior of the learned policies, extending the evaluation to more diverse tasks and exploring other non-explicitly reparameterizable distributions that could potentially be beneficial for injecting domain knowledge into the problem at hand. Additionally, more generic gradient estimators such as the score function estimator~\cite{estimators} could be investigated.


\bibliographystyle{abbrvnat}


\bibliography{bibliography}




\end{document}